\documentclass[10pt,twocolumn,letterpaper]{article}

\usepackage{cvpr}              
\usepackage{booktabs}
\usepackage{makecell}
\usepackage{subcaption}
\usepackage{graphicx}
\usepackage{colortbl}
\usepackage{microtype}
\usepackage{xcolor}
\usepackage{changepage}
\usepackage{svg}
\usepackage[export]{adjustbox}
\usepackage{stfloats}

\definecolor{cvprblue}{rgb}{0.21,0.49,0.74}
\usepackage[pagebackref,breaklinks,colorlinks,allcolors=cvprblue]{hyperref}

\def\paperID{37} 
\def\confName{ABAW}
\def\confYear{2026}

\title{Dimensional Distribution Emotion State: Leveraging Valence and Arousal as a Common Embedding Space for Visual Emotion Analysis}

\author{
    Émile Bergeron \qquad Tadagbé Dhossou \qquad Sébastien Tremblay \qquad Jean-François Lalonde \\
    Université Laval \\
}

\begin{document}

\newcommand{\heatmap}{\textsc{DDES}\xspace}
\newcommand{\catdist}{\textsc{CES}\xspace}
\newcommand{\vapoint}{\textsc{DES}\xspace}

\newcommand{\heatmapnet}{\textsc{\heatmap-Net}\xspace}
\newcommand{\catdistnet}{\textsc{\catdist-Net}\xspace}
\newcommand{\vapointnet}{\textsc{\vapoint-Net}\xspace}

\newcommand{\artemis}{ArtEmis\xspace}
\newcommand{\artelingo}{ArtELingo\xspace}
\newcommand{\dvisa}{D-ViSA\xspace}
\newcommand{\emoset}{EmoSet\xspace}
\newcommand{\eemobench}{EEmo-Bench\xspace}
\newcommand{\wikiartemotions}{WikiArt Emotions\xspace}

\maketitle
\begin{abstract}
\label{sec:abstract}
\noindent Museums are important sites for the dissemination of culture and art. They are institutions rooted in history and tradition; their exhibitions are often designed to highlight these aspects. Recently, a new approach is being explored in the field: emotion-based exhibitions. These exhibitions are designed specifically to elicit emotions in the visitors, in order to maximize engagement, and as a way to democratize access to art and attract a wider, more diverse audience. To do so, the emotional content of the artworks must first be extracted, however, manually annotating the artworks by experts is a prohibitively labor-intensive process, and risks introducing the personal bias of curators. To assist the museum curators in their design of these exhibitions, we wish to develop a tool that can predict the emotional response evoked by a work of art. In this article, we leverage a continuous bi-dimensional emotion space to enhance emotion representations and the training process of deep learning models. Drawing inspiration from existing categorical and dimensional emotion representations, we introduce a new representation, \textit{Dimensional Distribution Emotion State} (DDES), along with a pipeline for multi-dataset training. We show that DDES provides multiple advantages compared to widely used representations while exhibiting similar baseline performance. Project webpage: \href{https://github.com/echoes22/DDES}{https://github.com/echoes22/DDES}.

\end{abstract}

\section{Introduction}
\label{sec:intro}

Visual art is a powerful form of human expression; it is a medium favored by artists to convey strong meaning and emotion. In museums however, artworks are often displayed through an intellectual lens: they can be organized by genre, artist or historical period. 
In an effort to create more engagement and expose more people to visual art, museums have begun exploring the paradigm of emotion-centric exhibitions. 

A new challenge now arises: how can we know what emotions are most likely evoked by an artwork? Museum curators, while trained at describing and analyzing art, may not properly anticipate the emotional response of visitors since this is a highly subjective task. In the hope of helping curators get a better sense of emotional response to art, the computer vision community has explored the task of predicting emotions from images through the development of large datasets of images and associated emotions (\eg, \artemis~\cite{achlioptas2021artemis}), along with predictive models that can capture the real-world distribution of emotional responses (\eg, \cite{sevlm, circular-VAE}.

\begin{figure}[t!]
    \includegraphics[width=\columnwidth]{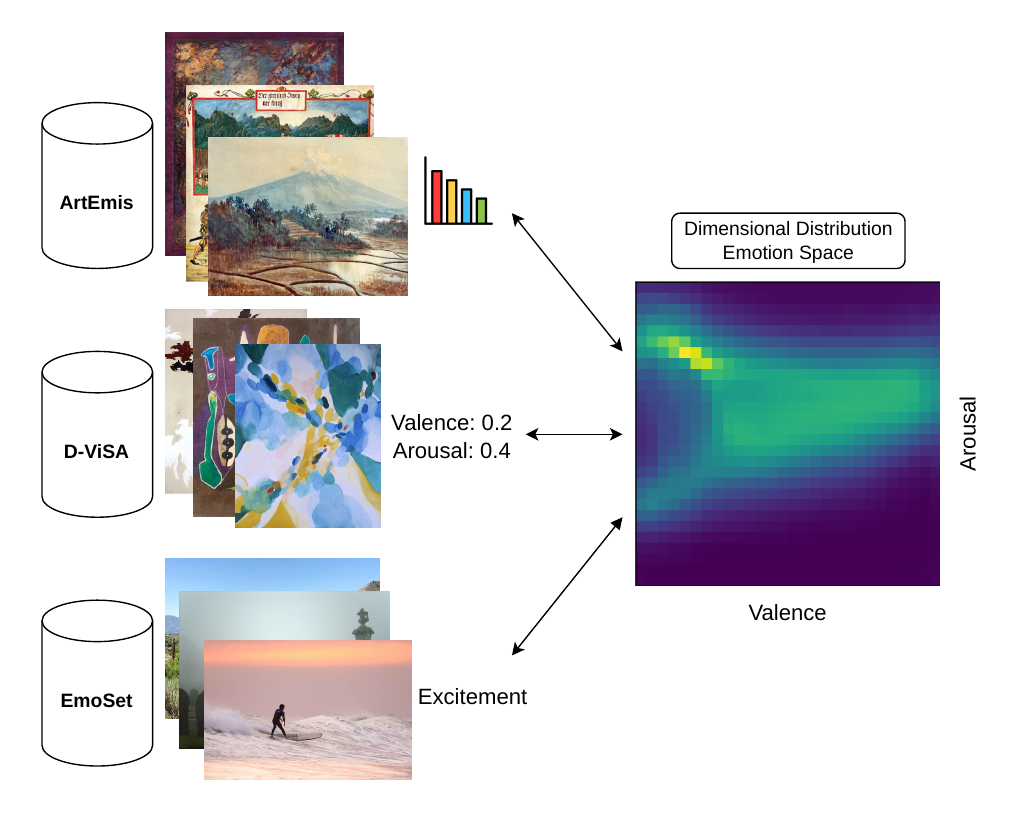}
    \caption{Our novel Dimensional Distribution Emotion Space (DDES) representation is used as a unifying space for datasets with different annotation modalities. It can be converted to Categorical Emotion Space (CES) and Dimensional Emotion Space (DES) through equations we define in this work, that leverage valence-arousal coordinates as mathematical grounding. The novel representation we propose has been designed for a concrete task: to assist museum curators in their artwork selection process for exhibitions. It addresses the issues with CES and DES representations, by simultaneously expressing nuanced and fine-grained emotions.}
    \label{fig:teaser}
\end{figure}


To study the applicability of such models in real museum exhibitions, we collaborated with curators and exhibition designers from the Musée national des beaux-arts du Québec (MNBAQ), who helped identify three desirable characteristics a predictive model should have. 
First, it should be capable of reflecting the nuance inherent to the human interpretation of art by allowing multiple emotions to be expressed simultaneously. Second, it should enable fine-grained details, by capturing as wide a spectrum of emotions as possible. Finally, the model should be able to leverage different training datasets, in order to fully take advantage of the limited data available. Unfortunately, current models and representations do not fulfill those requirements. Indeed, existing datasets contain labels in the form of distributions defined over a predefined, fixed set of emotions, which may vary across datasets, or in the form of a single numerical affective value. This then does not provide nuance over a wide range of emotions, and prevents models from being trained on multiple datasets.

In this paper, we propose a novel emotion representation, the Dimensional Distribution Emotion State (\heatmap), which can address these three limitations. By representing emotions as a probability distribution over the 2D space of valence and arousal from the psychology literature~\cite{russel, Valence-arousal, russel-integration}, this representation can express multiple, even opposing emotions simultaneously. It is not constrained to a discrete set of predefined emotions (\eg, Mikels wheel~\cite{mikels-wheel}), and it allows for multi-dataset training. 
This novel representation also has the advantage of easily aggregating various emotion labels---such as valence-arousal points, categorical labels and even full sentences---by projecting them to valence-arousal space. This allows future datasets following this method to conserve all of their crowd-sourced annotations, rather than consolidating them to a single value, allowing for richer datasets.

To assess the performance of our representation, we test it in a standard single dataset training scenarios, along with some zero-shot generalization benchmarks. We also compare it with two baseline models that use the two standard output representations. These baselines are also adapted with our conversion operations to allow for cross-representation training, to ensure fair comparisons. We find that the representation we introduce performs similarly to our baseline models on the training datasets, and generalizes better to unseen data in zero-shot scenarios.

\section{Related work}
\label{sec:related_works}

\subsection{The valence-arousal-dominance model}
\label{sec:valence-arousal}

In the field of psychology, the valence-arousal-dominance model \citep{russel, Valence-arousal, russel-integration} was developed to map emotions in a 3D space to serve as a tool for self-reporting emotion. The valence axis expresses how positive or negative an emotion is, the arousal axis indicates its intensity, and dominance indicates how in-control of the emotion the person is. For example, \textit{contentment}'s coordinates are [0.75;0.22;0.564], and \textit{sadness}'s are [-0.896;-0.424;-0.672]. Originally \citep{Valence-arousal}, a 9-point scale with pictograms for each dimension was used to collect the values from respondents.

Some interest has been manifested by the scientific community to find a reverse mapping, from words to VAD values \citep{Bradley1999AffectiveNF}. To this end, the NRC VAD lexicon~\cite{vad-acl2018} provides a dictionary of 20,000 words with their corresponding VAD values. It was created with crowd-sourced annotations, using the best-worse scaling method, where annotators are presented with 4-tuples, and then tasked with indicating the word with the best and the worst agreement with a particular axis. The aggregated rankings are then converted to fine-grained numerical values in the range of 0 to 1.

\subsection{Datasets}
There exists a corpus of datasets based on the artwork repository WikiArt, which includes 81k paintings. In \artemis \citep{achlioptas2021artemis}, \artemis V2 \citep{youssef2022artemis2} and \artelingo \citep{mohamed-etal-2022-artelingo}, the annotators were tasked with indicating the dominant emotion they feel when viewing each painting, along with a short affective explanation for their choice. These datasets are often used to train emotional speakers, \ie, models designed to provide emotional descriptions for works of art. A minimum of 5 annotations are provided for each painting, which allows the categorical labels to be aggregated per painting, thus creating an emotion distribution. They use emotion classification as an auxiliary task aimed at guiding the generation of the emotional caption. To do so, they focus on predicting the dominant emotion.

Some datasets with dimensional ground truth data are also publicly available. \dvisa \citep{dvisa} collects annotations for the valence, arousal and dominance axes, specifically for artworks in the genre of abstract expressionism. They also include a single dominant emotion label. The annotations are constructed by aggregating the response of 3 annotators, choosing a dominant emotion by majority, and averaging their dimensional values. 

\emoset \citep{EmoSet} uses an LLM to annotate 3.3 million images scraped from the web by keyword search. Their  dictionary of keywords is derived from Mikels 8 emotion model~\cite{mikels-wheel}, and serve as a ground truth for the collected image when they are in agreement with the model's prediction. They provide a subset of 118 thousand annotations that have been verified by humans. The annotations include the dominant emotion, along with dimensional values for valence and arousal.

\subsection{Emotion prediction}

Many methods have been developed to predict the affective content of artworks. Most of them focus on the task of predicting affective captions, and predict emotion label as an auxiliary task to inform the text generation \citep{achlioptas2021artemis, youssef2022artemis2}. 

Other works opt to depart from this task, and instead aim to predict different emotion representations with the goal of providing a richer emotional portrait. For instance, some focus on valence-arousal regression (dominance is often ignored as it is less expressive and harder to predict), in order to predict emotions without being limited to fixed coordinates. Zhang et al.~\citep{sevlm} train a Small Emotional Vision-Language model to jointly predict the dominant emotion, VAD vector and affective explanation using separate heads and a contrastive loss to align the results. They leverage the NRC VAD lexicon \citep{vad-acl2018} to enrich the textual features with a VAD vector computed from the input text. Other methods focus on the task of estimating valence and arousal based on different modalities of human expression, such as facial cues, voice, or brain activity \citep{Savchenko2023EmotiEffNetsFF, Wagner2022DawnOT, Galvao2021PredictingEV}.
In the field of natural language, \citep{dimensional-from-cat, va_text_pred} VA regression is also common. VA training objectives are obtained by mapping keywords to VA space. 

Other methods predict emotion distributions, in order to capture a more nuanced emotion response \citep{img-dist-learning, Xiong_Liu_Zhong_Fu_2019, Yang2017JointIE}. This tasked is called \textit{label distribution learning} (LDL). Yang et al. \citep{circular-VAE} arrange the 8 emotion labels in a circular structure, and they are given polar coordinates in order to apply an emotion informed circular loss. Lee et al. \citep{enhancing-dim} train a model to jointly regress a VA value and a categorical distribution to enhance performance on dimensional datasets with limited data. Some early work \citep{Zhao2017ContinuousPD} has attempted to predict a distribution in VA space, by algorithmically regressing gaussian mixture model parameters, but little follow up work has been done. 
In contrast, our method combines both paradigms of distribution-based and dimensional-based emotion prediction methods, by mapping aggregated annotations to VA space, creating a novel joint distribution regression task that can be framed as \textit{dimensional distribution learning} (DDL). 

\section{Numerical representations for emotions}
\label{sec:emo_reps}

We begin by describing various ways of representing emotions numerically, that can be amenable to predicting using deep neural networks. All are based on the well-known valence-arousal model from the psychology literature (c.f., \cref{sec:valence-arousal}). These representations are either classified as Categorical Emotion States (CES), or Dimensional Emotion States (DES).  We also present how to map to-and-from each of the representations. In practice, while the standard VAD model includes the additional dimension \textit{dominance}, it is generally not included in datasets, as its meaning is harder to communicate to annotators. Consequently, we also exclude it from our work.


\subsection{Three emotion representations}

\paragraph{The Categorical Emotion State (\catdist)} is a probability distribution over a set of $N$ discrete emotions. For example, one could use Mikels' wheel model~\citep{mikels-wheel}: \textit{awe, amusement, contentment, excitement, disgust, sadness, fear, anger}, where $N=8$. The probability of each emotion is stored in a vector $\mathbf{p} \in \mathbb{R}^N$. Each emotion can also be associated with a 2D valence-arousal coordinate $\mathbf{c}_i \in \mathbb{R}^2$ according to the NRC lexicon~\cite{vad-acl2018}.
While this representation allows the expression of nuance, by enabling weight to be given to multiple, possibly opposing emotions, it is limited to a fixed number of emotion. Datasets that use this representation include \artemis \citep{achlioptas2021artemis}, \artemis V2 \citep{youssef2022artemis2}, \artelingo \citep{mohamed-etal-2022-artelingo}, Affection \citep{Achlioptas2022AffectionLA}, EMOTIC \citep{Kosti2017EMOTICEI} and \wikiartemotions \citep{LREC18-ArtEmo}.


\paragraph{The Dimensional Emotion State (\vapoint)} simply consists of a single point $\mathbf{v} \in \mathbb{R}^2$ in the 2D valence-arousal space. While this representation is flexible and could potentially express \emph{any} emotion (not just those restricted to a list, as in the previous case), it can only represent a single one at a time. Datasets that use this representation include \dvisa~\citep{dvisa},
\eemobench~\citep{eemo-bench}, EmoArt~\citep{emoart}, FindingEmo~\citep{findingemo} and EMOTIC~\citep{Kosti2017EMOTICEI}. 


\paragraph{The Dimensional Distribution Emotion State (\heatmap)} is the joint 2D distribution of emotions expressed in the 2D valence-arousal space. In practice, we store a discretized version of the distribution in a 2D matrix $\mathbf{Z} \in \mathbb{R}^{H \times W}$. This representation is nuanced, as it allows multiple regions of the space to be activated. It is also fine-grained, because it can express any emotion defined in VA space. 



\begin{figure}
    \includegraphics[width=\linewidth]{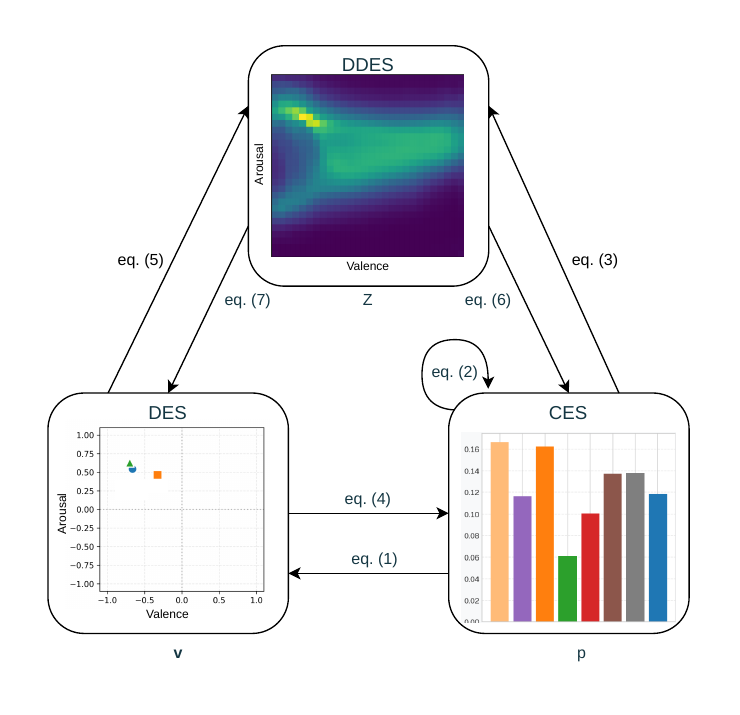}
    \caption{Conversion operations are defined using the intermediary VA space. We can go from \heatmap to a categorical distribution or VA point using \cref{eq:heatmap2catdist} and \cref{eq:heatmap2vapoint} respectively. To convert from a VA point to categorical distribution, we use \cref{eq:vapoint2catdist}, and \cref{eq:vapoint2heatmap} for \heatmap. To map a \catdist to another \catdist, we use \cref{eq:catdist2catdist}. To convert to \vapoint and \heatmap, we use \cref{eq:catdist2vapoint} and \cref{eq:catdist2heatmap}, respectively.}
    \label{fig:va_conversion}
\end{figure}

\subsection{Operations for representation conversion}
\label{sec:conversion}

We now define conversion operations to map from each emotion representation to the others. The conversions are summarized in \cref{fig:va_conversion}; all exploit the common 2D valence-arousal space. 



\subsubsection{From \catdist}

\paragraph{To \vapoint.} A single VA point $\mathbf{v}$ can be obtained via a weighted average of the emotion labels:
\begin{equation} \mathbf{v} = \sum_{i=1}^{N} p_i \mathbf{c}_i \,, 
\label{eq:catdist2vapoint} 
\end{equation}
where $N$ is the total number of emotion classes in the distribution. 


\paragraph{To a different \catdist.} To map a categorical distribution $\{(p_i, \mathbf{c}_i)\}_{i=1}^N$ to a different one (containing different emotion labels) $\{(q_i, \mathbf{d}_i)\}_{i=1}^M$, we compute the distances in VA space from each of the target emotions to the source emotions, and compute the resulting probabilities as the product of the source probabilities and the inverse of the distance to target distribution as follows:
\begin{equation} 
    q_i = \frac{\sum_{j=1}^{N} \frac{p_j}{|\mathbf{d}_i - \mathbf{c}_j| + \epsilon}}{\sum_{k=1}^M \sum_{j=1}^N \frac{p_j}{|\mathbf{d}_k - \mathbf{c}_j| + \epsilon}}  \,,
    \label{eq:catdist2catdist} 
\end{equation}
where $\epsilon$ is a stability constant. Note that here, the weights are calculated using inverse distance, rather than gaussian softmax, in order to preserve the influence of further emotional anchors, to ensure a smooth distribution when the labels are distributed sparsely in VA space.

\paragraph{To \heatmap.} A gaussian kernel is applied to each coordinate of the input emotion set $\mathbf{c}_k$ and weighted by their respective probabilities $p_k$:
\begin{equation}
    \mathbf{Z}_{i,j} = \sum_{k=1}^{N} p_k \exp\left( -\frac{\| \mathbf{z}_{i,j} - \mathbf{c}_k \|^2}{2\sigma^2} \right) \,,
    \label{eq:catdist2heatmap}
\end{equation}
where $x_{i,j}$ are the coordinates of grid cell at row $i$, column $j$. Variance $\sigma$ controls the width of each gaussian kernel. In practice, the covariance is computed using Scott's Rule \citep{Scott1992MultivariateDE}.

\subsubsection{From \vapoint}

\paragraph{To \catdist.} 
The euclidean distance from the VA point to each emotion center is first computed, then normalized to obtain the probability distribution over the desired set of emotions:
\begin{equation} 
    \mathbf{p}_i = \frac{e^{-k |\mathbf{v} - \mathbf{c}_i|^2}}{\sum_{j=1}^{N} e^{-k |\mathbf{v} - \mathbf{c}_j|^2}} \,,
\label{eq:vapoint2catdist} 
\end{equation}
where $k$ is a scaling factor controlling the sharpness of the distribution. In practice, we use $k=10$.

\paragraph{To \heatmap.} We again apply a gaussian kernel as in \cref{eq:catdist2heatmap}, but in this case, for only one unweighted point:

\begin{equation}
    \mathbf{Z}_{i,j} = \exp\left( -\frac{\| \mathbf{z}_{i,j} - \mathbf{v} \|^2}{2\sigma^2} \right) \,.
    \label{eq:vapoint2heatmap}
\end{equation}

\subsubsection{From \heatmap}

\paragraph{To \catdist.}
The probability $p_i$ at each emotion coordinate $\mathbf{c_i}$ is sampled from $\mathbf{Z}$ using bilinear interpolation:
\begin{equation} 
    \mathbf{p}_i = \text{Bilinear}(\mathbf{Z}, \mathbf{c}_i) \,.
    \label{eq:heatmap2catdist}
\end{equation}

\paragraph{To \vapoint.} 
We first apply a softmax to the input probability distribution $\mathbf{Z}$:
\begin{equation} 
    \tilde{\mathbf{Z}}_i = \frac{(\mathbf{Z}_i + \epsilon)^{1/\tau}}{\sum_{i=1}^{WH} (\mathbf{Z}_i + \epsilon)^{1/\tau}} \,,
    \label{eq:heatmap2vapoint}
\end{equation}
where the temperature $\tau$ can be adjusted to obtain a sharper or blurrier distribution.  Here, we use $\tau = 0.05$. The resulting VA point $\mathbf{v}$ is the center of mass of $\tilde{\mathbf{Z}}$. 




\begin{figure*}[t]
    \includegraphics[width=\linewidth]{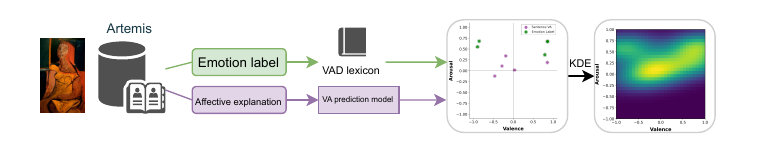}
    \caption{Constructing \heatmap annotations for the ``\artemis Enhanced'' dataset. The original annotations from \artemis (emotion labels and affective explanations) are first processed into a point cloud using the VAD lexicon and a VA prediction model. Then, a kernel density estimation is applied, and the density function is subsequently sampled to obtain the emotional density grid.}
    \label{fig:kde_visualisation}
\end{figure*}

\section{Experiments}
\label{sec:experiments}

In our experiments, we define three models, each using one of the three representations described in \cref{sec:emo_reps}, and subject them to different training and evaluation scenarios. Namely, we compare the effectiveness of training on single datasets (\cref{sec:single-dataset-training}) versus training jointly on multiple datasets (\cref{sec:multi-dataset-training}). In both cases, we evaluate the models on in-distribution and out-of-distribution data. 

\subsection{Models}

For each model, we adapt a pretrained ConvNeXt \cite{liu2022convnet} backbone with a custom decoder to output the desired representation.



\paragraph{\catdistnet.} The output layer is linear with $N$ output features, one for each emotion in the discrete distribution (which can vary across datasets), and add a log-softmax layer to output a distribution in log space. 

\paragraph{\vapointnet.} The output layer is linear with 2 output features, one for the valence dimension, and one for arousal.

\paragraph{\heatmapnet.} We create a custom decoder based on the ConvNext architecture that takes as input the output of the final layer (\textit{features.7}), and spatially upsamples the features, while progressively downsampling the channels. The spatial resolution is upsampled twice by a factor of two in the two decoder blocks, before undergoing a final $1 \times 1$ convolution, reducing the feature channels to 1, with a final resolution of $28 \times 28$. A log softmax layer is applied at the end, to obtain the log-probabilities of the density function at every cell.

\paragraph{Pre-trained VLM.} We extract the emotional understanding of a VLM in three different ways, mirroring the three emotion representations of the previous models. All the implementation details are in the supplementary, \cref{sec:vlm_representations}.

\subsection{Datasets}
\label{subsec:datasets}

\paragraph{\artemis.} 

We combine both \artemis and \artemis V2, for a total of more than 700k annotations. Recall from \cref{subsec:datasets} that annotations include both discrete dominant emotions and short affective explanations. To enhance the dataset beyond the $N=8$ emotions, we convert each emotion label and explanation sentence to a single valence-arousal point using the NRC VAD lexicon~\cite{vad-acl2018} and a pretrained language model~\cite{va_text_pred} respectively. The set of resulting VA points is then converted to a \heatmap representation using \cref{eq:vapoint2heatmap} (adapted to process multiple VA points). The process is summarily illustrated in \cref{fig:kde_visualisation}. We name this resulting dataset ``\artemis Enhanced'' (referred to as \artemis from now on).

\paragraph{\dvisa.} 
We simply apply a linear scaling to the provided \vapoint labels, bringing them to the [-1, 1] range. We apply no further pre-treatment to the dataset. We also opt to not use the categorical labels during training, as \dvisa is one of the only datasets built specifically to collect valence and arousal values. 

\paragraph{\emoset.} \emoset is composed of real-world images, scraped from many online sources and separated by emotion class (following Mikel's 8 emotion model) using keywords. Six types attributes describing the scene are generated using a VLM (such as scene type, colorfulness or human facial expression), accompanying the emotion label. Specifically, we use \emoset-118k, a human-verified subset of the machine annotated larger scale datasets of 3.3M annotations. In this subset, training examples are only chosen if 7/10 human annotators are in agreement for the different model-predicted attributes. The resulting emotion label can thus be seen as an approximately unimodal distribution.

\paragraph{\wikiartemotions.}
\wikiartemotions selects approximately 200 artworks for each of the 22 art categories in the Wikiart image bank, for a total of 4105 images. For each image, they collect 10 crowd-sourced annotations, by asking the annotator to select the dominant emotion across a set of $N=20$ options (see supp. \cref{full_emo_sets}). They offer different versions of the dataset, by filtering out samples where the mass of the dominant emotion is below certain thresholds. We use the unfiltered version, with all the samples.

\paragraph{\eemobench.} \eemobench is made from images scraped from Flickr, and grouped by emotion using keywords. A set of $N=7$ emotion is used (see supp. \cref{full_emo_sets}). For each image, 15 annotators are tasked with choosing 3 emotions and ranking them by strength, and choosing valence, arousal and dominance scores from the SAM 9-point model \citep{Bradley1999AffectiveNF}. The top-3 emotion are then computed using a ranking algorithm, and the VAD scores by averaging the middle 9 scores from the 15 total annotations.

\subsection{Single dataset training}
\label{sec:single-dataset-training}

This first set of experiments aims to evaluate the performance of the different models on the specific dataset they were trained on. The aim is to assess how the added complexity of the dimensional distribution affects the performance for downstream tasks like classification or valence-arousal regression. Quantitative results are reported in \cref{tab:per-dataset}. 

\paragraph{ArtEmis training.} This dataset contains \heatmap ground truth labels. We therefore train each model by converting the \heatmap label to their respective representation using \cref{eq:heatmap2vapoint} for \vapointnet, and \cref{eq:heatmap2catdist} for \catdistnet (see \cref{sec:conversion}). For evaluation, we chose to measure top-1 accuracy, F1 score and Kendall's tau ranking correlation ($\tau$) \citep{KendallsTau} on the  \catdist inferred from the ground truth \heatmap, in order to obtain interpretable metrics. A Kullback-Leibler loss is used for both \catdistnet and \heatmapnet  models, as their log-probability output is tailored for this function, and for the \vapointnet model, a Mean Squared Error loss (MSE) is used.

\paragraph{\dvisa training.} 
When training on \dvisa, we only use the valence-arousal points as ground truth labels, and convert the output of each model to VA vectors using \cref{eq:catdist2vapoint} for \catdistnet and \cref{eq:heatmap2vapoint} for \heatmapnet (see \cref{sec:conversion}). We use an MSE loss to train every model. The performance is then evaluated on the test set, by calculating the Pearson correlation \citep{Pearson01061931} separately on both dimensions.

\paragraph{\emoset training.}
\emoset only provides one categorical label per artwork. To convert this to a representation usable by our 3 networks, we map the emotion labels to their VA coordinates from the NRC VAD lexicon, and use this as a training objective, in a similar fashion as the training for \dvisa. Then, at test time, we convert each model's output to \catdist and measure top-1 accuracy and F1 score with the original categorical labels, like in the \artemis training.



\paragraph{Zero-shot generalization.}
To assess generalization capabilities, we evaluate every model, trained on only one dataset, on each of the other datasets. As before, the conversion functions from \cref{sec:conversion} are used to adapt each the output of each model to the target emotion representation. Since the datasets contain a variety of representations and domains (artworks, abstract paintings, real-world photos), this allows us to assess performance on 0-shot generalization across domain and representation of the different models. \Cref{tab:cross_ds_gen} introduces the results of the performance of \catdistnet, \vapointnet and \heatmapnet on this scenario. 


\subsection{Multi-dataset training}
\label{sec:multi-dataset-training}

Leveraging the joint 2D valence-arousal space, we combine training examples from different datasets. To this end, examples are sampled from all three datasets (\artemis, \dvisa, \emoset) in a training batch, and the appropriate losses, as described in \cref{sec:single-dataset-training} are employed for each type of ground truth data. The loss is then scaled by the proportion of training examples of this type in the batch.

\begin{table}[t]
\centering
\caption{Comparison of models trained using the three different emotion representations (\catdist, \vapoint, and \heatmap). The models are trained and evaluated on every dataset separately.}
\label{tab:per-dataset}
\resizebox{\columnwidth}{!}{%
    \begin{tabular}{@{}l ccc ccc cc@{}}
    \toprule
    \textbf{Model} & 
    \multicolumn{3}{c}{\textbf{\artemis}} & 
    \multicolumn{3}{c}{\textbf{\dvisa}} & 
    \multicolumn{2}{c}{\textbf{\emoset}} \\
    
    \cmidrule(lr){2-4} 
    \cmidrule(lr){5-7} 
    \cmidrule(lr){8-9} 
    
     & Acc$_\uparrow$ & F1$_\uparrow$ & $\tau_\uparrow$
     & $r_{v\uparrow}$ & $r_{a\uparrow}$ & RMSE$_\downarrow$ 
     & Acc$_\uparrow$ & F1$_\uparrow$ \\
     
    \midrule
    \catdistnet & \underline{39.7} & \textbf{17.2} & \underline{0.32} & 0.445 & \textbf{0.139} & \textbf{0.31} & \textbf{71.1} & \textbf{50.9} \\
    \vapointnet  & 29.4 & 11.5 & 0.21 & \textbf{0.501} & \underline{0.092} & \underline{0.32} & \underline{60.0} & \underline{45.0} \\
    \heatmapnet   & \textbf{39.8} & \underline{15.9} & \textbf{0.34} & \underline{0.450} & 0.081 & 0.33 & 50.5 & 39.3 \\
    \bottomrule
    \end{tabular}
}
\end{table}

\begin{table*}[t]
    \centering
    \caption{Cross-dataset generalization results. The highlighted cells indicate cases where the train and test set are the same, corresponding to the reference \cref{tab:per-dataset}.}
    \label{tab:cross_ds_gen}
    
    \begin{subtable}{0.32\linewidth}
        \centering
        \caption{Training Set: \textbf{\artemis}}
        \resizebox{\linewidth}{!}{%
            \begin{tabular}{l c cc c}
                \toprule
                & \multicolumn{4}{c}{\textbf{Target (Eval) Dataset}} \\
                \cmidrule(lr){2-5}
                \textbf{Model} & 
                \textbf{Art.} & \multicolumn{2}{c}{\textbf{\dvisa}} & \textbf{Emo.} \\
                \cmidrule(lr){3-4}
                 & \small{(Acc)} & \small{($r_v$)} & \small{($r_a$)} & \small{(Acc)} \\
                \midrule
                \catdistnet & \cellcolor{yellow!25}\underline{39.7} & \textbf{0.369} & 0.012 & \textbf{22.6} \\
                \vapointnet  & \cellcolor{yellow!25}29.4 & 0.290 & \textbf{0.055} & 17.4 \\
                \heatmapnet   & \cellcolor{yellow!25}\textbf{39.8} & \underline{0.295} & \underline{0.038} & \underline{19.4} \\
                \bottomrule
            \end{tabular}
        }
    \end{subtable}%
    \hfill
    \begin{subtable}{0.32\linewidth}
        \centering
        \caption{Training Set: \textbf{\dvisa}}
        \resizebox{\linewidth}{!}{%
            \begin{tabular}{l c cc c}
                \toprule
                & \multicolumn{4}{c}{\textbf{Target (Eval) Dataset}} \\
                \cmidrule(lr){2-5}
                \textbf{Model} & 
                \textbf{Art.} & \multicolumn{2}{c}{\textbf{\dvisa}} & \textbf{Emo.} \\
                \cmidrule(lr){3-4}
                 & \small{(Acc)} & \small{($r_v$)} & \small{($r_a$)} & \small{(Acc)} \\
                \midrule
                \catdistnet & 9.3  & \cellcolor{yellow!25}\underline{0.445} & \cellcolor{yellow!25}\textbf{0.139} & \underline{16.5} \\
                \vapointnet  & \underline{17.0} & \cellcolor{yellow!25}\textbf{0.501} & \cellcolor{yellow!25}\underline{0.092} & \textbf{19.3} \\
                \heatmapnet   & \textbf{21.4} & \cellcolor{yellow!25}0.450 & \cellcolor{yellow!25}0.081 & 15.7 \\
                \bottomrule
            \end{tabular}
        }
    \end{subtable}%
    \hfill
    \begin{subtable}{0.32\linewidth}
        \centering
        \caption{Training Set: \textbf{\emoset}}
        \resizebox{\linewidth}{!}{%
            \begin{tabular}{l c cc c}
                \toprule
                & \multicolumn{4}{c}{\textbf{Target (Eval) Dataset}} \\
                \cmidrule(lr){2-5}
                \textbf{Model} & 
                \textbf{Art.} & \multicolumn{2}{c}{\textbf{\dvisa}} & \textbf{Emo.} \\
                \cmidrule(lr){3-4}
                 & \small{(Acc)} & \small{($r_v$)} & \small{($r_a$)} & \small{(Acc)} \\
                \midrule
                \catdistnet & \textbf{15.6} & \underline{0.213} & -0.016 & \cellcolor{yellow!25}\textbf{71.1} \\
                \vapointnet  & 12.0 & \textbf{0.226} & \textbf{0.096}  & \cellcolor{yellow!25}\underline{60.0} \\
                \heatmapnet   & \underline{14.7} & 0.187 & \underline{0.075}  & \cellcolor{yellow!25}50.5 \\
                \bottomrule
            \end{tabular}
        }
    \end{subtable}
\end{table*}



\paragraph{Additional benchmark datasets.}
We use two additional datasets to evaluate generalization: no training is performed on them. The first is \eemobench~\citep{eemo-bench}, which consists of 1960 real-world images and is designed to assess the performance of VLMs on varied tasks. We only use the provided top-3 emotions and valence-arousal values in the annotations. This dataset employs a set of 7 emotions, differing from the standard established by \artemis. The second test set is \wikiartemotions \citep{LREC18-ArtEmo}, composed of 4105 artworks selected across 4 western art styles from the WikiArt image bank. This time, a set of 20 emotions is used to collect the annotations. To evaluate \catdistnet on these different sets of emotions, we resample its 8-emotion output using \cref{eq:catdist2catdist}.

\begin{table}[t]
\centering
\setlength{\tabcolsep}{2pt}
\caption{Comparison of models trained jointly on \artemis, \dvisa and \emoset, evaluated on the datasets seen at training time, and two unseen benchmark datasets.}
\label{tab:multi-ds-generalization}
\resizebox{\columnwidth}{!}{%
    \begin{tabular}{@{}l cc ccc c ccc cc@{}}
    \toprule
    & \multicolumn{6}{c}{\textbf{Seen Datasets}} 
    & \multicolumn{5}{c}{\textbf{Unseen Datasets}} \\
    \cmidrule(lr){2-7} \cmidrule(lr){8-12}

    \textbf{Model} & 
    \multicolumn{2}{c}{\textbf{\artemis}} & 
    \multicolumn{3}{c}{\textbf{\dvisa}} & 
    \multicolumn{1}{c}{\textbf{\emoset}} & 
    \multicolumn{3}{c}{\textbf{\eemobench}} & 
    \multicolumn{2}{c}{\textbf{Wiki. Emo.}} \\
    
    \cmidrule(lr){2-3} 
    \cmidrule(lr){4-6} 
    \cmidrule(lr){7-7} 
    \cmidrule(lr){8-10} 
    \cmidrule(lr){11-12}
    
     & Acc$_\uparrow$ & $\tau_\uparrow$
     & $r_{v\uparrow}$ & $r_{a\uparrow}$ & RMSE$_\downarrow$ 
     & Acc$_\uparrow$ 
     & $r_{v\uparrow}$ & $r_{a\uparrow}$ & Acc$_\uparrow$ 
     & Acc$_\uparrow$ & $\tau_\uparrow$  \\
     
    \midrule
    \catdistnet    & \textbf{38.9} & \textbf{0.32} & \underline{0.42} & 0.05 & 0.61 & \textbf{60.2} & \textbf{0.59} & 0.12 & \underline{18.9} & \underline{8.6} & 0.07 \\
    \vapointnet    & 24.1 & 0.19 & 0.38 & \underline{0.06} & \underline{0.60} & \underline{50.1} & \underline{0.58} & \underline{0.23} & 15.8 & 7.6 & \underline{0.13} \\
    \heatmapnet & \underline{33.6} & \underline{0.30} & \textbf{0.43} & \textbf{0.09} & \textbf{0.55} & 39.4 & 0.47 & \textbf{0.32} & \textbf{21.2} & \textbf{10.5} & \textbf{0.18} \\
    \bottomrule
    \end{tabular}
}
\end{table}

\subsection{Observations}
\label{sec:observations}
The results observed in \cref{tab:per-dataset,tab:cross_ds_gen,tab:multi-ds-generalization} allow us to make a number of observations that can be summarized as follows.

\paragraph{Models perform best when trained solely on the dataset on which they are evaluated.} The highest performance for a single dataset are always observed when the model is trained on that same dataset (\cref{tab:per-dataset}). 
The highest accuracy on \artemis is 39.8\% is obtained with \heatmapnet and closely followed by \catdistnet at 39.7\%. On \dvisa, the best Pearson score for the valence dimension is \vapointnet at 0.501, and for arousal, it is 0.139 with the \catdistnet model. Finally, on \emoset, the highest accuracy is 71.1\% with \catdistnet. This is not surprising since there is no other signal in the training process; the model therefore best adapts to a particular dataset.

\paragraph{Models perform best when their output representation matches the evaluation dataset representation.} The datasets used for training and evaluation all use categorical labels or valence-arousal points as their emotion representation. It can be observed in \cref{tab:cross_ds_gen} that regardless of the training set, \vapointnet performs best on \dvisa which uses VA points, while \catdistnet performs best on \artemis and \emoset, which both use categorical labels for evaluation. Considering results across all of \cref{tab:per-dataset,tab:cross_ds_gen}, we observe that \vapointnet exhibits the best performance on \dvisa in 3 of the 6 cases, but never on the categorical datasets. Conversely, \catdistnet is the best performer in 4 of the 6 scenarios on the categorical datasets. \heatmapnet appears to be less sensitive to different representations, displaying more stable performance regardless of the evaluation set representation. A similar trend can be observed in \cref{tab:multi-ds-generalization}, where the best scores on \artemis and \emoset are obtained with \catdistnet, while \vapointnet performs best on \dvisa and the VA regression part of \eemobench, and is far behind the other models in categorical tasks.


\paragraph{The continuous valence-arousal space can successfully be used to convert output representations.} While it has been established that models perform best when their output representation matches the dataset, we observe that the continuous valence-arousal space provides a consistent semantic grounding for mapping between different emotion representations. In \cref{tab:cross_ds_gen}, models trained on \artemis and \emoset (both having categorical \catdist labels) can still successfully transfer their visual emotion analysis capabilities to \dvisa to perform zero-shot valence-arousal regression, obtaining positive correlations for both valence and arousal. 

\paragraph{\heatmapnet performs best on unseen datasets and unseen representations.}  \Cref{tab:multi-ds-generalization} compares the models on unseen datasets, which contain both new images and emotion representations as labels. Recall from \cref{subsec:datasets} that the training datasets \artemis and \emoset use the same 8 emotions, while the unseen datasets \eemobench and \wikiartemotions use \emph{different} ones (7 for \eemobench and 20 for \wikiartemotions). In this scenario, we observe that \heatmapnet achieves the best performance across all but one metrics, showing its capacity of generalizing to completely unseen emotion representations at test time, in a zero-shot manner. 


\paragraph{Multi-dataset training lowers the ceiling but increases the floor for performance across all seen datasets.} When training jointly on all 3 datasets, we observe in \cref{tab:cross_ds_gen,tab:multi-ds-generalization} that the maximum performance on a dataset is always lower than what is obtained when training on a single dataset. On \artemis, the difference in accuracy is relatively small (38.9\% vs 39.7\% accuracy). A larger gap can be observed with Pearson correlations  on \dvisa, at 0.43 vs 0.501 and 0.09 vs 0.139 for valence and arousal resp. Similarly, on \emoset, the accuracy drops significantly, at 60.2\% vs 71.1\%. However, models never reach the extremely low performance exhibited in the cross-dataset generalization scenario, having seen all three datasets at train time.

\begin{figure*}[t]
    \centering
    \setlength{\tabcolsep}{1pt} 
    \renewcommand{\arraystretch}{0.5} 
    
    \begin{tabular}{cccc}
        \textbf{Artwork} & \textbf{\heatmap} & \textbf{\catdist} & \textbf{\vapoint} \\[2mm]
        
        \includegraphics[height=2.2cm]{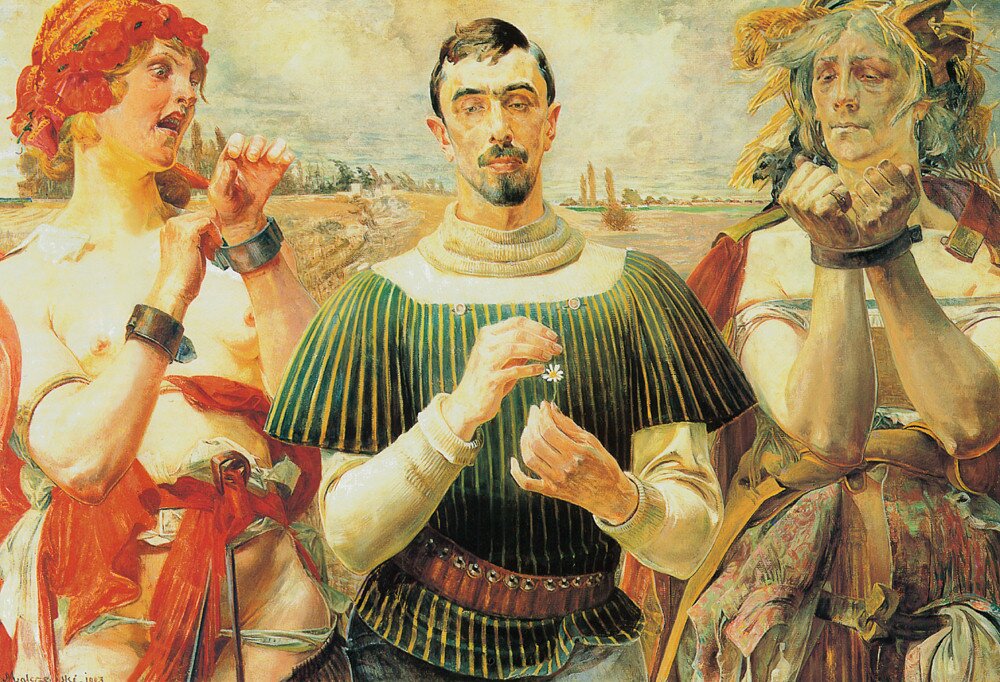} &
        \includegraphics[height=2.2cm]{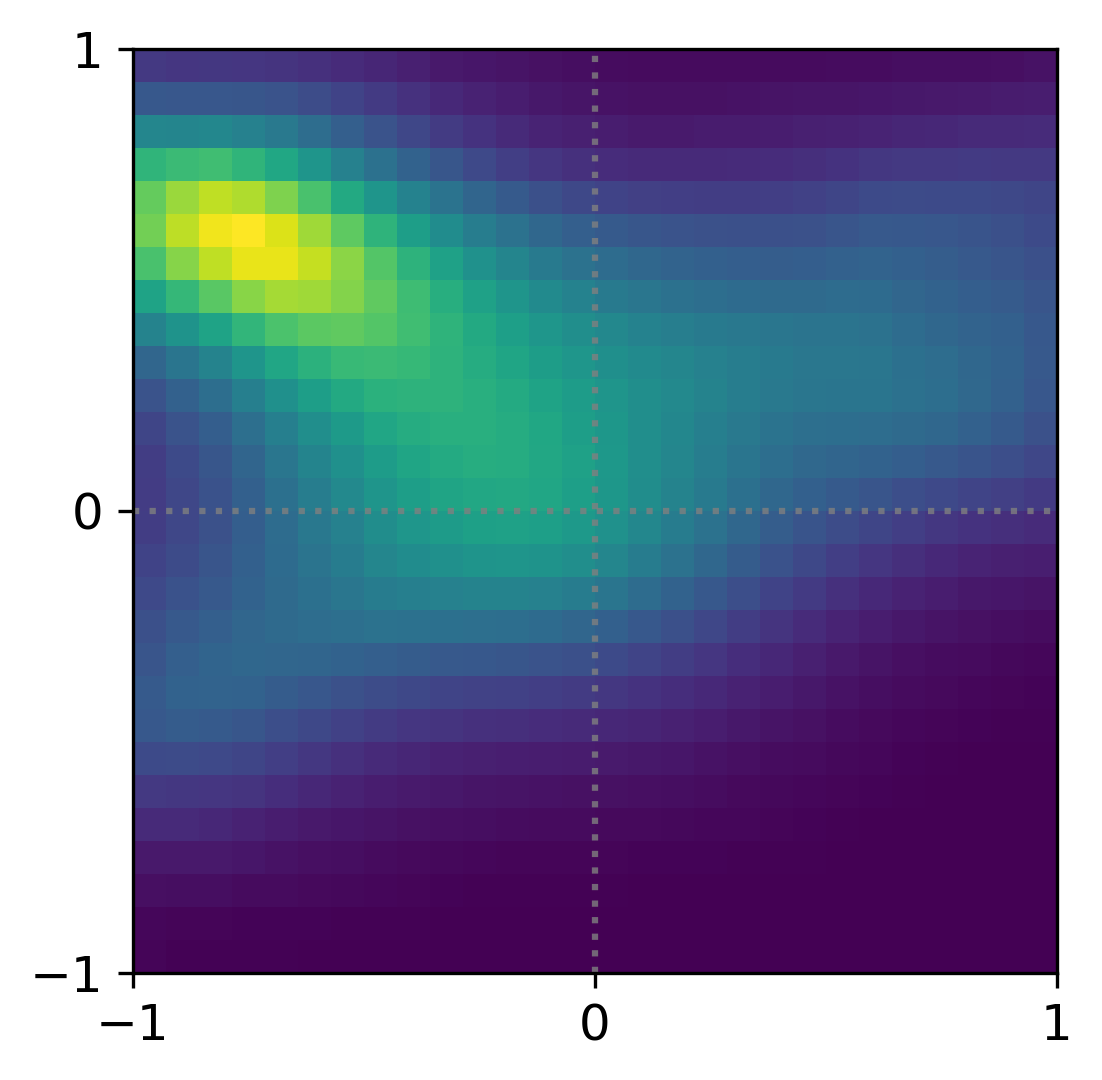} &
        \includegraphics[height=2.2cm]{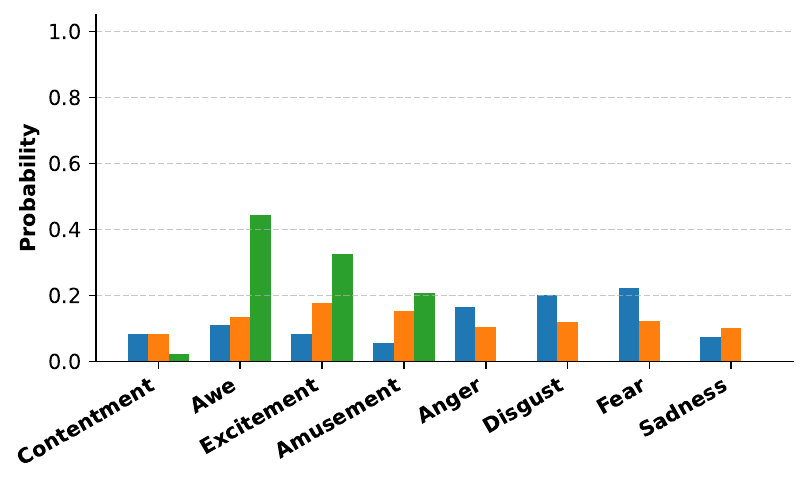} &
        \includegraphics[height=2.2cm]{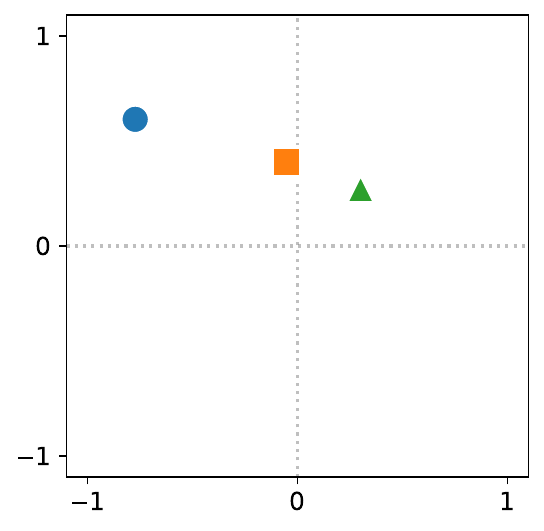} \\
        \scriptsize \textit{Portrait of Aleksander Wielopolski} & & & \\[2mm]
        
        \includegraphics[height=2.2cm]{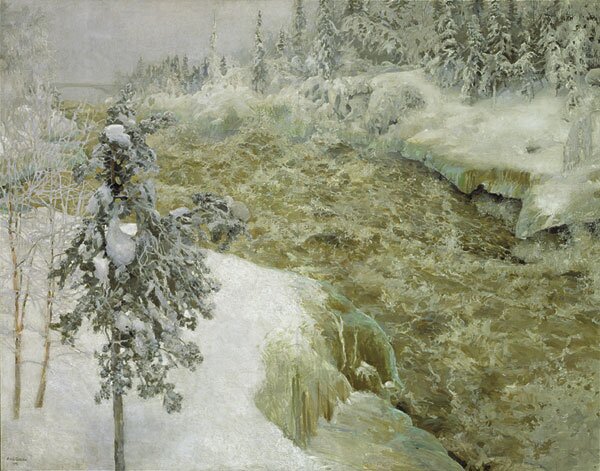} &
        \includegraphics[height=2.2cm]{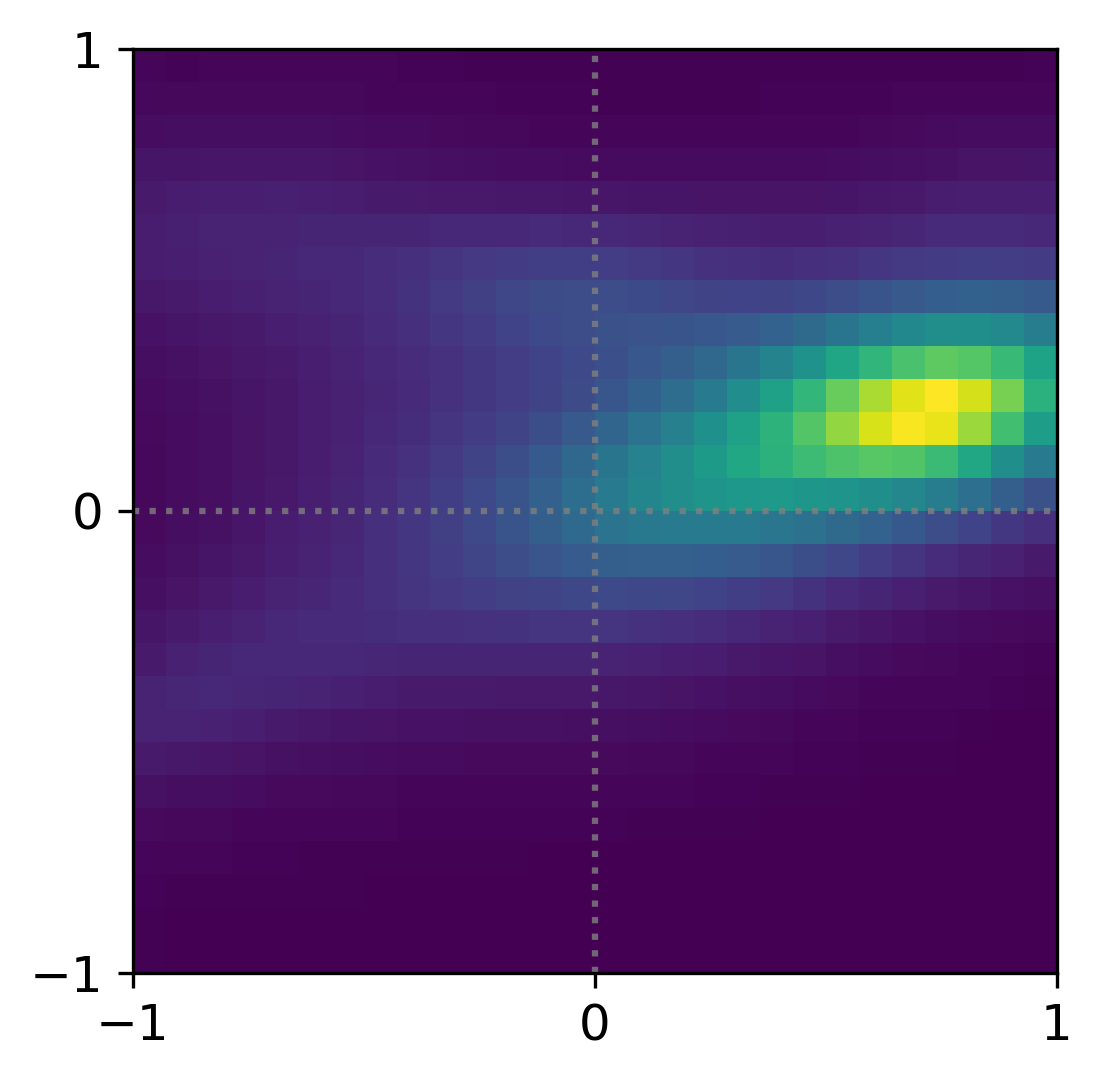} &
        \includegraphics[height=2.2cm]{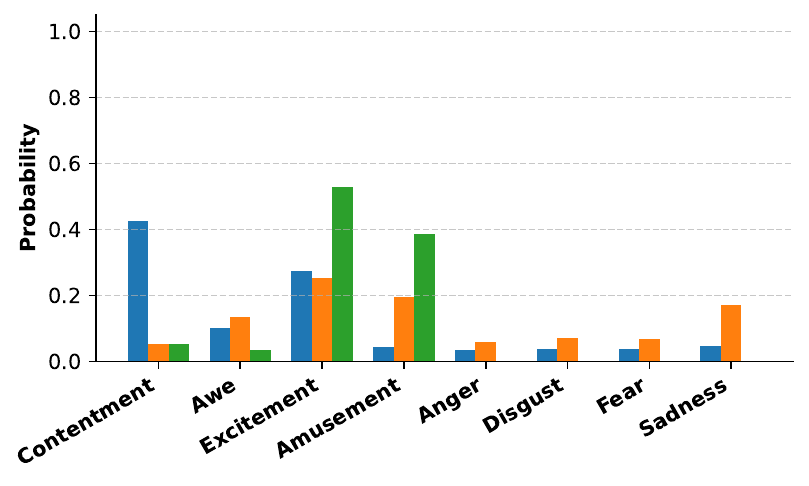} &
        \includegraphics[height=2.2cm]{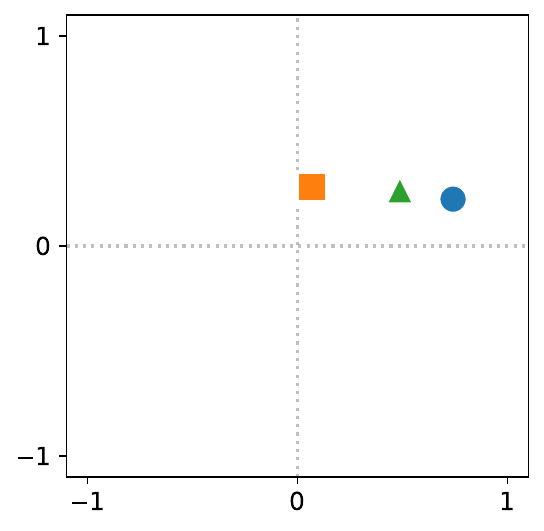} \\
        \scriptsize \textit{Imatra in Winter} & & & \\[2mm]

        \includegraphics[height=2.2cm]{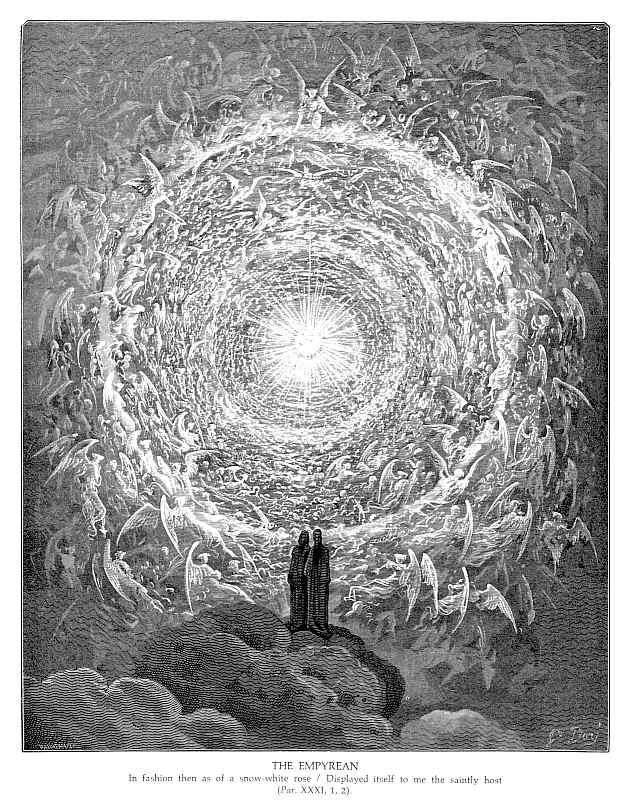} &
        \includegraphics[height=2.2cm]{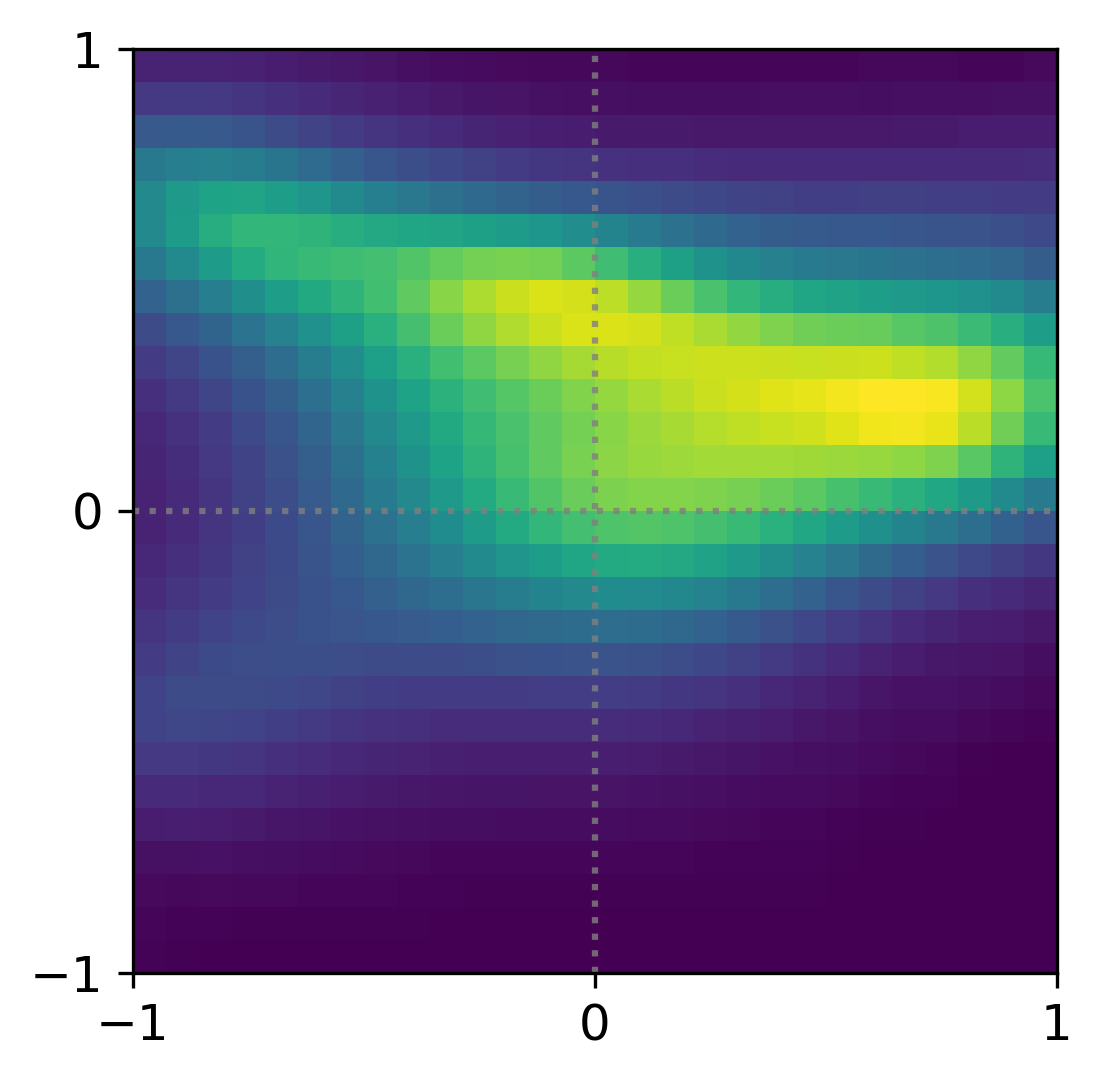} &
        \includegraphics[height=2.2cm]{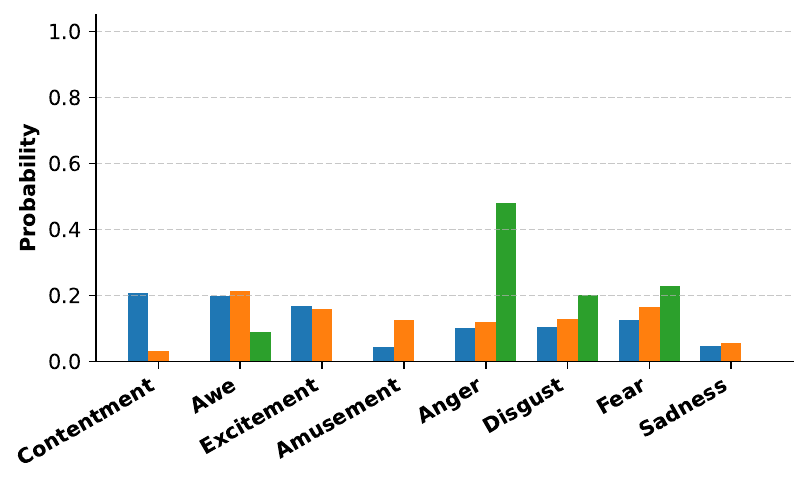} &
        \includegraphics[height=2.2cm]{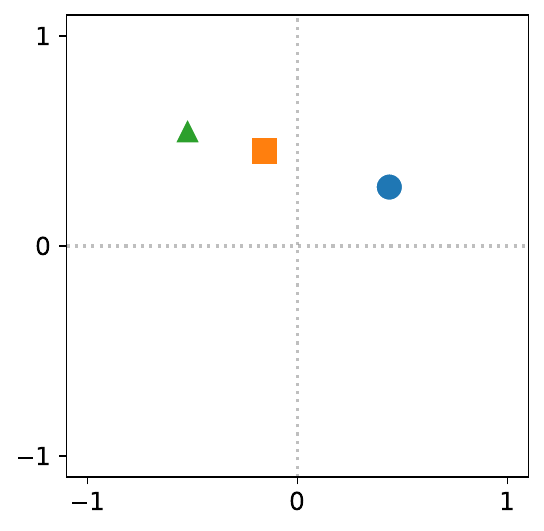} \\
        \scriptsize \textit{The Empyrean} & & & \\[2mm]
        
        \multicolumn{4}{c}{\includegraphics[width=0.5\linewidth]{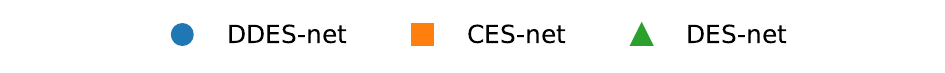}}
    \end{tabular}
    
    \caption{Visualization of predictions from the three different models \heatmapnet, \vapointnet, \catdistnet. In the two right-most columns, the predictions are converted to the column's representation using the conversion equations from \cref{sec:conversion}.}
    \label{fig:qual_results}
\end{figure*}

\begin{figure}[ht]
    \centering
    \begin{subfigure}[b]{0.32\linewidth}
        \centering
        \includegraphics[height=3cm]{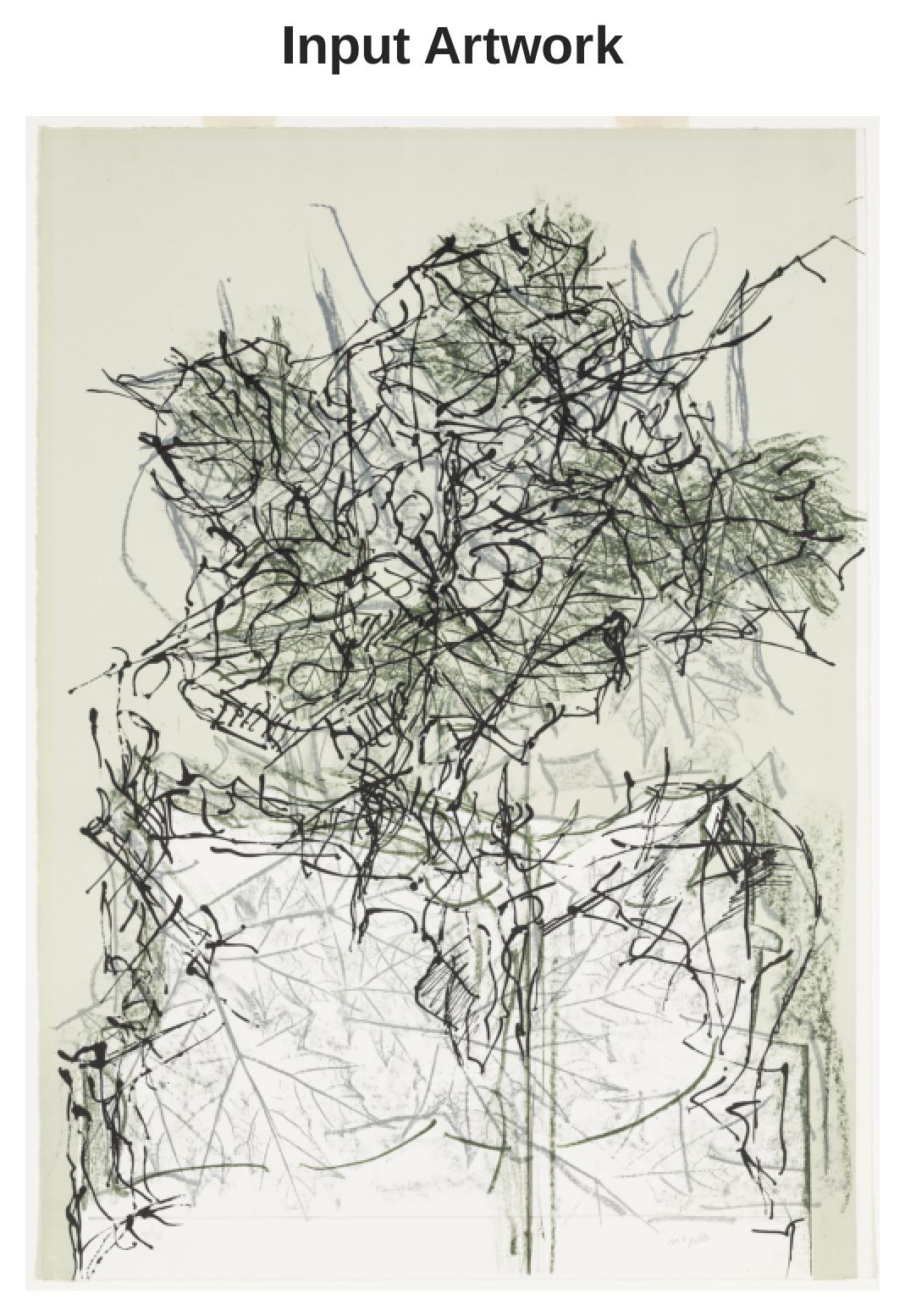}
        \caption{}
        \label{fig:sub_artwork}
    \end{subfigure}\hfill
    \begin{subfigure}[b]{0.32\linewidth}
        \centering
        \includegraphics[width=\linewidth]{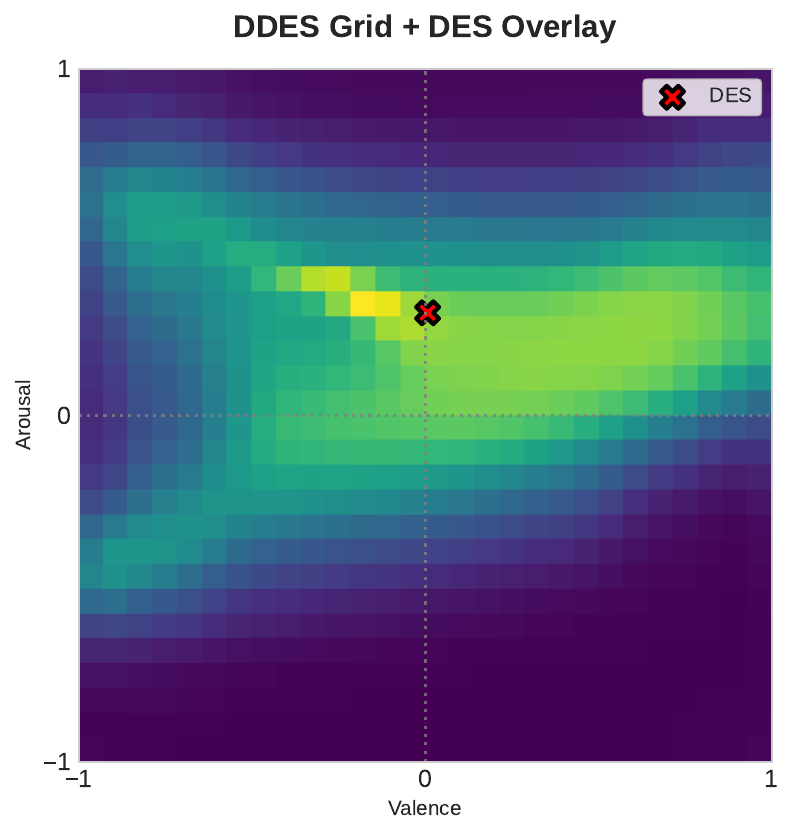}
        \caption{}
        \label{fig:sub_heatmap}
    \end{subfigure}\hfill
    \begin{subfigure}[b]{0.32\linewidth}
        \centering
        \includegraphics[width=\linewidth]{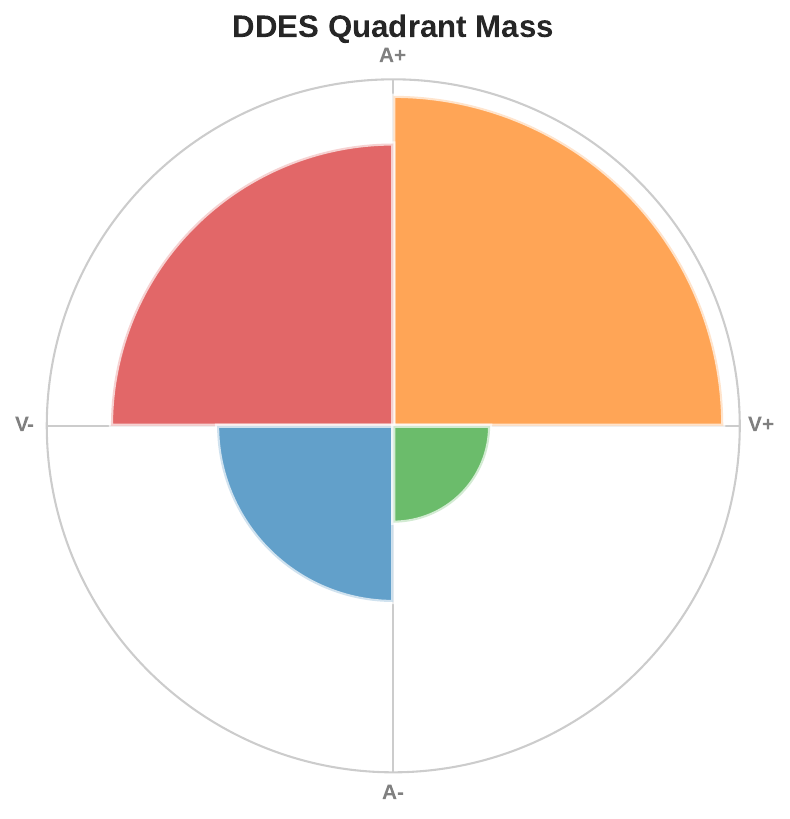}
        \caption{}
        \label{fig:sub_rose}
    \end{subfigure}
    
    \vspace{2mm} 
    
    \begin{subfigure}[b]{0.32\linewidth}
        \centering
        \includegraphics[width=\linewidth]{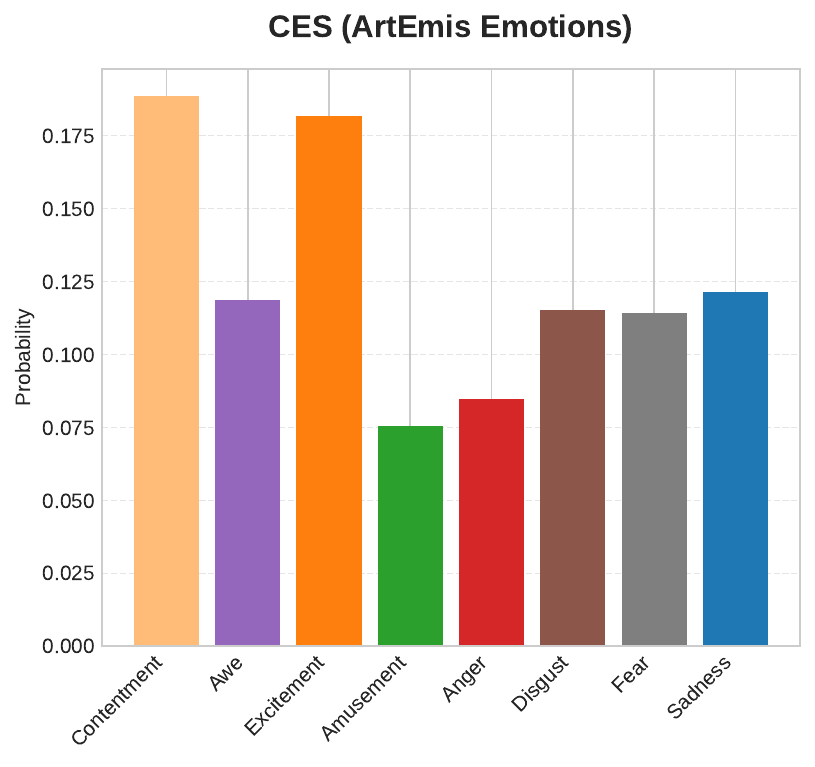}
        \caption{}
        \label{fig:sub_ces_base}
    \end{subfigure}\hfill
    \begin{subfigure}[b]{0.32\linewidth}
        \centering
        \includegraphics[width=\linewidth]{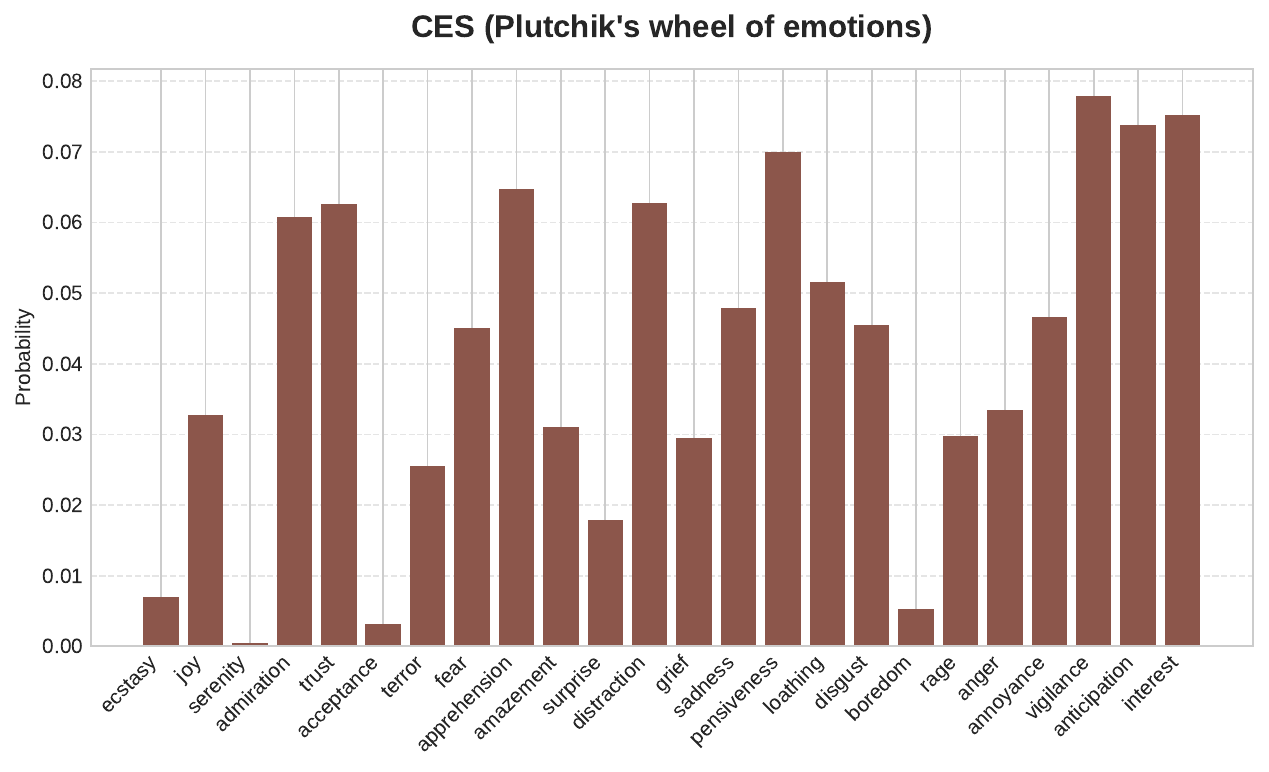}
        \caption{}
        \label{fig:sub_ces_alt}
    \end{subfigure}\hfill
    \begin{subfigure}[b]{0.32\linewidth}
        \centering
        \includegraphics[width=\linewidth]{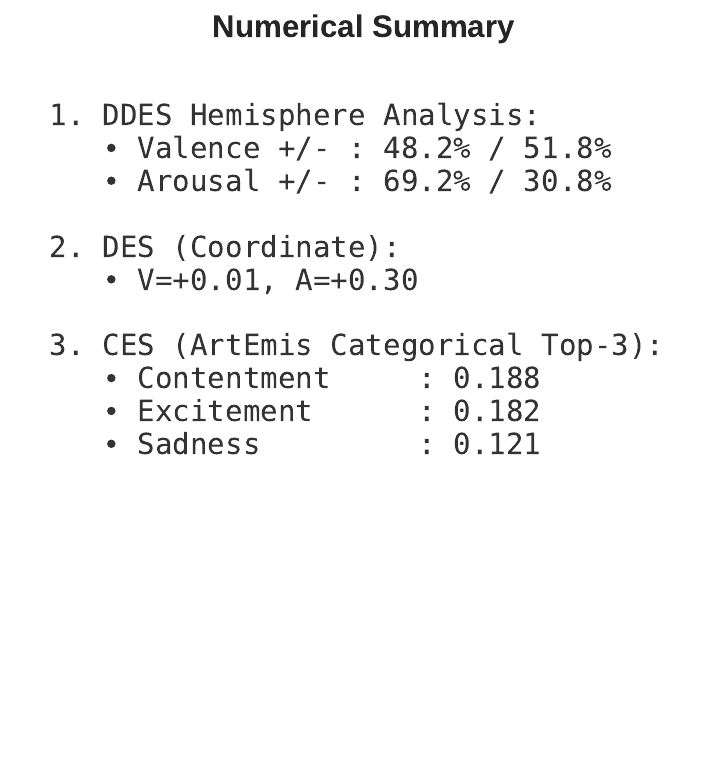}
        \caption{}
        \label{fig:sub_summary}
    \end{subfigure}
    
    \caption{Example of \heatmap analysis. It can be easily projected to the \catdist in \artemis (d) or in \wikiartemotions (e), or to a \vapoint (b). The \heatmap can also be analyzed for coarser-grained information, like mass per quadrant (c) or mass per hemisphere (f).}
    \label{fig:heatmap_analysis}
\end{figure}

\paragraph{\artemis Enhanced significantly increases generalization performance.} 
In \cref{tab:multi-ds-supp} (see \cref{sec:additional_results} of the supplementary document), we observe  that performance drops when the combined training is done with the original version of \artemis. We speculate that incorporating the VA points of the descriptive sentences introduces mass in other areas in the VA space than the fixed coordinates of the emotion set. This gives the model the ability to predict mass in an unconstrained way, leading to much better performance when the evaluation set is not in the exact coordinate system used during training. We do however observe a caveat; the model tends to not allocate mass in the lower right quadrant (see \cref{fig:qual_results,fig:heatmap_analysis}), as a result of the data distribution in \artemis, which limits its generalization to emotion sets defined in this area of the VA space.



\subsection{Generating other visualizations}

The \heatmap representation allows for the generation of additional emotion visualizations, even if they are not explicitly present in the ground truth labels. This is illustrated in \cref{fig:qual_results}, where it is sampled to obtain a distribution on Plutchik's wheel of emotions \citep{theNatureOfEmotions}, a more fine-grained emotion model with 27 different categories. The repartition of the mass in the \heatmap can also be visually analyzed by computing the proportion of the mass per quadrant, or per hemisphere.



\section{Conclusion}
\label{sec:conclusion}

Our objective was to develop a novel method to help guide the design process of museum exhibitions by providing rich emotional annotations. In this paper, we introduce a new paradigm in visual emotion analysis: leveraging the valence-arousal space defined in psychological studies to provide an emotional grounding for different representations. Using this space, we introduce a novel emotion representation, the Dimensional Distribution Emotion State or \heatmap, a joint 2D valence-arousal distribution. Conversion operations can be used to convert between different emotion representations, and we contrast our \heatmap with two commonly-used representations: \catdist (distribution over a fixed set of emotions) and \vapoint (valence-arousal point). We train models to predict each one, and assess their performance with standard supervised training scenarios and with zero-shot generalization on benchmark datasets with differing emotion representations. We find that generally, while models perform best on datasets that match their output representation, \heatmapnet is able to adapt to learn both \catdist and \vapoint representations, and generalizes best to datasets with emotion sets not seen during training, while providing a richer basis for emotional analysis. We also show that the multi-dataset training pipeline leads to more robust models that have a better understanding of the different representations. 
\\
While we explored training on multiple datasets simultaneously, we did not use every publicly available affective dataset. Notably, we chose not to include \artelingo, an extension of \artemis with annotations in four languages, as it would require an additional translation step that could potentially introduce a semantical bias. We also decided not to include other affective datasets with non-artistic images, such as Affection \citep{Achlioptas2022AffectionLA} or EMOTIC \citep{Kosti2017EMOTICEI}, as we wanted to primarily focus on artworks; however, they might improve performance as a form of transfer learning. We also acknowledge that the predictions of the valence-arousal estimation model incorporated in the \artemis ground truth may not perfectly reflect the annotators sentiment. 
Future research could focus on the network architecture side. Notably, diffusion models could be investigated as a potential feature extraction backbone through inversion. Their innate understanding of art could provide a strong prior for visual emotion analysis.


\paragraph*{Acknowledgements}

This research was partially supported by NSERC, Musée National des Beaux-Arts du Québec and Strateolab, as well as FRQ-NT and NSERC MS fellowships to E. Bergeron. Computing resources were provided by the Digital Research Alliance Canada. The authors thank our lab members for help with proofreading. 

{
    \small
    \bibliographystyle{ieeenat_fullname}
    \bibliography{main}
}

\clearpage
\onecolumn
\setcounter{page}{1}
\maketitlesupplementary

\section{Adapting large pre-trained VLMs for emotion prediction} 
In the main section of this paper, we train models specifically for the task of predicting the emotional response evoked by artworks. In this section, drawing inspiration from EEmo-Bench \citep{eemo-bench}, our goal is to determine whether large pretrained VLMs can be adapted for predicting the different emotion representations identified in this paper without retraining. 

\subsection{VLM adaptation}
\label{sec:vlm_representations}
We devise different ways of mapping the model's internal representation to the three emotion representations from \cref{sec:emo_reps}. To do so, we extract the predicted probabilities of the next token given a context prompt using the following methods.

\paragraph{\catdist representation}
To obtain a \catdist representation, we give the model a contextual prompt: ``Assume you are an expert in emotional psychology. Choose one word from the following list: (\texttt{<emotion-list-str>}). Which emotion is primarily expressed in this image? The primary emotion is''. The model then chooses the most likely token in its vocabulary to start its answer. At this moment, we extract the probabilities of each emotion label using their precomputed token IDs, before undergoing a normalization to obtain the final distribution. The list of emotions \texttt{<emotion-list-str>} is defined according to the emotion labels present in the corresponding dataset. We sometimes run into the case where a single emotion label is split into multiple tokens by the tokenizer. To circumvent this issue, we establish a mapping from single-token, rarely used symbols to our target emotion set, and place it into the prompt.

\paragraph{\vapoint representation}
To obtain a \vapoint representation, we follow the same methodology as \citep{eemo-bench}, where we prompt the model with the following text for each of valence/arousal: ``How would you rate the valence/arousal this image evokes in the viewer? The level of valence/arousal this image evokes in the viewer is''. The query tokens are ``Negative'', ``Neutral'', ``Positive'', and  probabilities for each token are then weighted with [-1.0, 0, 1.0] respectively, before being normalized using softmax.

\paragraph{\heatmap representation}
To obtain a \heatmap representation, we first identified some emotional keywords from the NRC VAD lexicon \citep{vad-acl2018} such that their valence-arousal coordinates spanned the 2D space as thoroughly and equally as possible. Each symbol is mapped to its corresponding keyword in the following prompt: ``Assume you are an expert in emotional psychology. Analyze the viewer's emotion evoked by this image. The emotion codes are mapped as follows: \textit{$\Diamond$ is boredom, $\circledcirc$ is hopeless, $\triangle$ is unhappy, $\triangledown$ is humiliated, $\spadesuit$ is rage, $\heartsuit$ is shy, $\diamondsuit$ is timid, $\clubsuit$ is beholden, $\bigstar$ is perplexed, $\star$ is flustered, $\blacklozenge$ is sleepy, $\lozenge$ is contemplation, $\bullet$ is reverent, $\circ$ is forceful, $\square$ is fanatical, $\blacksquare$ is mellow, $\blacktriangle$ is reflective, $\blacktriangledown$ is yearning, \dag\ is expectant, \ddag\ is elation, \S\ is calming, \P\ is grace, \textcurrency\ is prized, \texteuro\ is proud, \pounds\ is surprise}. Which single code from the set (\texttt{<token-labels>}) best represents the highest probability emotion? The code is''  \\
Then, we extract the probabilities for every keyword, and compute a KDE on the resulting weighted point cloud. The density function is then discretized and normalized, following our method. This is perhaps not the emotion representation most adapted to a VLM's representation, but we wanted to include it for comparison purposes.

\subsection{Experimental results}
In all of our experiments, we use the open-source Qwen-2.5-7B VLM model. We performed the same experiments as in \cref{sec:experiments}. Note that we could not however replicate the seen/unseen scenarios, as we have to assume that a model of this scale has seen \artemis, \dvisa, \emoset, \eemobench and \wikiartemotions during training.

\begin{table*}[t]
\centering
\caption{Comparison of different methods of extracting emotional comprehension from VLMs (QWEN-2.5-7B).}
\label{tab:vlm}
\resizebox{\textwidth}{!}{%
    \begin{tabular}{@{}l cc ccc c ccc cc@{}}
    \toprule
    \textbf{Mapping Method} & 
    \multicolumn{2}{c}{\textbf{\artemis}} & 
    \multicolumn{3}{c}{\textbf{\dvisa}} & 
    \multicolumn{1}{c}{\textbf{\emoset}} & 
    \multicolumn{3}{c}{\textbf{\eemobench}} & 
    \multicolumn{2}{c}{\textbf{\wikiartemotions}} \\
    
    \cmidrule(lr){2-3} 
    \cmidrule(lr){4-6} 
    \cmidrule(lr){7-7} 
    \cmidrule(lr){8-10} 
    \cmidrule(lr){11-12}
    
     & Acc$_\uparrow$ & $\tau_\uparrow$
     & $r_v$$_\uparrow$ & $r_a$$_\uparrow$ & RMSE$_\downarrow$ 
     & Acc$_\uparrow$ 
     & $r_v$$_\uparrow$ & $r_a$$_\uparrow$ & Acc$_\uparrow$ 
     & Acc$_\uparrow$ & $\tau_\uparrow$ \\
     
    \midrule
    \quad Logits to \catdist & \textbf{39.9} & \underline{0.138} & \textbf{0.477} & \underline{0.009} & \underline{0.648} & \textbf{49.8} & \underline{0.750} & \textbf{0.421} & \textbf{52.1} & \textbf{34.2} & \underline{0.121} \\
    \quad Logits to \vapoint           & \underline{16.3} & \textbf{0.249} & \underline{0.325} & \textbf{0.105} & \textbf{0.644} & 19.7 & 0.682 & 0.358 & 19.9 & \underline{10.9} & \textbf{0.127} \\
    \quad Logits to \heatmap      & 15.4 & -0.008 & 0.297 & -0.067 & 0.870 & \underline{24.4} & \textbf{0.757} & \underline{0.400} & \underline{21.1} & 2.83 & -0.144 \\
    \bottomrule
    \end{tabular}

}
\end{table*}

\subsection{Observations}

\paragraph{Querying a VLM leads to similar performance as supervised training on seen datasets.} When compared the models trained on the combined datasets in \cref{tab:multi-ds-generalization}, we can observe in \cref{tab:vlm} that the VLM leads to very similar performance on the seen datasets. Qwen performs slightly better than the best supervised model on \artemis and \dvisa, and considerably worse on \emoset. 

\paragraph{VLMs perform much better on benchmark datasets, but can not be considered zero-shot generalization.}On the benchmark datasets, Qwen considerably outperforms the supervised models in every metric. On \eemobench, the difference is stark, with 0.757 against 0.59 Pearson correlation on the valence axis, and 0.421 against 0.32 on the arousal axis. The same applies for accuracy, at 52.1 against 21.2. Again, on \wikiartemotions, we observe a considerable gap with 34.2 versus 10.5 for top-1 accuracy. This is a much bigger difference in performance than for the three seen datasets, which is to be expected as it is not a fair comparison. The VLM has seen the datasets during its training, and not only the test splits, but the train splits as well. The very large amount of data it has ingested prevents it from remembering every training example, but it still gives it an innate advantage over the specialized models.

\paragraph{Categorical emotion states are the representation most suited to extract a VLM's emotional understanding.} The logits to \catdist method yields the best results for most metrics in \cref{tab:vlm}; it is best or second best in all cases. This representation is also the one that most closely matches the usual inference process of a VLM, which is based on predicting probabilities for text tokens.

\section{Additional results}
\label{sec:additional_results}

\subsection{Experiments with \heatmapnet training}
Additional experiments with \heatmapnet under the single-dataset training and combined training settings outlined in \cref{sec:experiments}. The results are reported in \cref{tab:multi-ds-supp}. These experiments serve to explore research directions orthogonal to the choice of emotion representation.

\paragraph{Single-dataset} We assess \heatmapnet's zero-shot generalization capabilities by testing it on our two benchmark datasets, \eemobench and \wikiartemotions. We find that the model trained solely on \artemis actually significantly outperforms the model trained on the combined datasets when looking at top-1 accuracy for both datasets, but exhibits lower performance for valence-arousal regression on \eemobench.

\paragraph{Multi-dataset} We test \heatmapnet on the combined training setting, with 3 variations. The first one adds an auxiliary loss inspired by \citep{circular-VAE}. Here, the predictions and ground truth are first converted to \vapoint representation using our equations from \cref{sec:conversion}, and a position, norm and orientation loss are computed on the resulting vectors. These loss terms are then simply added to the original loss. This additional loss slightly increases performance on some metrics, while decreasing it on others. We chose not to use this as baseline.
Other loss variations could be explored, such as Earth Mover's Distance (EMD), or regularization terms could be added, for instance to penalize more heavily zones with no mass in the ground truth. 

The balanced sampling experiment consists of sampling each dataset an equal amount of time during training. In practice, this heavily upsamples \dvisa, which is a much smaller size than \artemis and \emoset. This leads to slightly increased performance on \dvisa, \artemis and \wikiartemotions, but much lower performance on \emoset and \eemobench. If the combined training procedure is expanded, and more affective datasets are added to the pool, sampling might become a more important concern, but for now didn't prove to be necessary.

Lastly, we substituted the enhanced \artemis with the base version in the combined training pipeline. We observe drastically worse results, as reported in \cref{sec:observations}


\begin{table*}[t]
\centering
\caption{Additional experiments conducted on the \heatmapnet model. In the single dataset training scenario, we evaluate a \heatmapnet model trained solely on \artemis on the additional benchmark datasets, \eemobench and \wikiartemotions. In the combined training section, we experiment with adding an auxiliary loss, sampling the datasets samples equally each epoch instead of proportionally to their size, and with using the base version of \artemis that does not incorporate the emotion points from the affective explanations.}
\label{tab:multi-ds-supp}
\resizebox{\textwidth}{!}{%
    \begin{tabular}{@{}l cc ccc c ccc cc@{}}
    \toprule
    & \multicolumn{6}{c}{\textbf{Seen Datasets}} 
    & \multicolumn{5}{c}{\textbf{Unseen Datasets}} \\
    \cmidrule(lr){2-7} \cmidrule(lr){8-12}

    \textbf{Experiment} & 
    \multicolumn{2}{c}{\textbf{\artemis}} & 
    \multicolumn{3}{c}{\textbf{\dvisa}} & 
    \multicolumn{1}{c}{\textbf{\emoset}} & 
    \multicolumn{3}{c}{\textbf{\eemobench}} & 
    \multicolumn{2}{c}{\textbf{\wikiartemotions}} \\
    
    \cmidrule(lr){2-3} 
    \cmidrule(lr){4-6} 
    \cmidrule(lr){7-7} 
    \cmidrule(lr){8-10} 
    \cmidrule(lr){11-12}
    
     & Acc$_\uparrow$ & $\tau_\uparrow$
     & $r_{v\uparrow}$ & $r_{a\uparrow}$ & RMSE$_\downarrow$ 
     & Acc$_\uparrow$ 
     & $r_{v\uparrow}$ & $r_{a\uparrow}$ & Acc$_\uparrow$ 
     & Acc$_\uparrow$ & $\tau_\uparrow$  \\
     
    \midrule
    
    \multicolumn{12}{@{}l}{\textbf{Single-Dataset Training}} \\
    \midrule
    \artemis only & 40.3 & 0.34 & 0.34 & -0.03 & 0.58 & 18.6 & 0.35 & 0.18 & 31.7 & 13.2 & 0.21 \\
    
    \midrule
    \multicolumn{12}{@{}l}{\textbf{Combined Training}} \\
    \midrule
    Auxiliary loss & 39.1 & 0.30 & 0.37 & 0.06 & 0.53 & 48.1 & 0.55 & 0.13 & 26.8 & 9.8 & 0.19 \\
    Balanced sampling & 40.1 & 0.25 & 0.47 & 0.10 & 0.45 & 14.0 & 0.38 & 0.06 & 6.7 & 10.0  & 0.16 \\
    Base \artemis & 13.9 & 0.17 & 0.184 & -0.058 & 0.49 & 15.1 & 0.299 & 0.058 & 17.1 & 6.8  & 0.136 \\
    
    \bottomrule
    \end{tabular}
}
\end{table*}

\section{Implementation details}
\subsection{Full emotion sets}
\label{full_emo_sets}
The following are the full emotion sets used in datasets mentioned in \cref{subsec:datasets}.

\paragraph{\eemobench} The full set of emotions is: \textit{joy, surprise, fear, disgust, sadness, anger and neutral}.

\paragraph{\wikiartemotions} The full set of emotions is: \textit{agreeableness, anger, anticipation, arrogance, disagreeableness,disgust, fear, gratitude, happiness,
 humility, love, optimism, pessimism,
 regret, sadness, shame, shyness,
 surprise, trust, and neutral}.

\end{document}